\documentclass[letterpaper]{article} % DO NOT CHANGE THIS
\usepackage{aaai2026}  % DO NOT CHANGE THIS
\usepackage{times}  % DO NOT CHANGE THIS
\usepackage{helvet}  % DO NOT CHANGE THIS
\usepackage{courier}  % DO NOT CHANGE THIS
\usepackage[hyphens]{url}  % DO NOT CHANGE THIS
\usepackage{graphicx} % DO NOT CHANGE THIS
\usepackage{algorithm}
\usepackage{algorithmic}
\usepackage{multirow}
\usepackage{booktabs}  % For better table formatting
\usepackage{multirow}
\usepackage{tabularx}
\usepackage{arydshln}
\usepackage{graphicx}

\usepackage{latexsym}
\usepackage{colortbl}
\usepackage[table]{xcolor}

\usepackage{epsfig}
\usepackage{amsmath}
\usepackage{amssymb}
\usepackage{booktabs}
\usepackage{subcaption}
\usepackage{breqn}
\usepackage{pgfplots}
\usepackage{adjustbox}
\usepackage{dsfont}
\pgfplotsset{compat=1.16}

\usepackage{pifont}
\newcommand{\cmark}{\ding{51}}%
\newcommand{\xmark}{\ding{55}}%
\newcommand{\model}{QDETRv}

\usepackage{natbib}  % DO NOT CHANGE THIS AND DO NOT ADD ANY OPTIONS TO IT
\usepackage{caption} % DO NOT CHANGE THIS AND DO NOT ADD ANY OPTIONS TO IT
\usepackage{algorithm}
\usepackage{algorithmic}

\usepackage{newfloat}
\usepackage{listings}
\DeclareCaptionStyle{ruled}{labelfont=normalfont,labelsep=colon,strut=off} % DO NOT CHANGE THIS
\floatstyle{ruled}
\newfloat{listing}{tb}{lst}{}
\floatname{listing}{Listing}
\title{Temporal Object-Aware Vision Transformer for \\Few-Shot Video Object Detection}
\author{
    Yogesh Kumar and Anand Mishra\\
}
\affiliations{
    Indian Institute of Technology Jodhpur, India
    kumar.204@iitj.ac.in, mishra@iitj.ac.in
}

\usepackage{bibentry}
\begin{document}

\maketitle

\begin{abstract}
Few-shot Video Object Detection (\textsc{fsvod}) addresses the challenge of detecting novel objects in videos with limited labeled examples, overcoming the constraints of traditional detection methods that require extensive training data. This task presents key challenges, including maintaining temporal consistency across frames affected by occlusion and appearance variations, and achieving novel object generalization without relying on complex region proposals. Our novel object-aware temporal modeling approach addresses these challenges by incorporating a filtering mechanism that selectively propagates high-confidence object features across frames. This enables efficient feature progression, reduces noise accumulation, and enhances detection accuracy in a few-shot setting. By utilizing few-shot trained detection and classification heads with focused feature propagation, we achieve robust temporal consistency without depending on explicit object tube proposals. Our approach achieves performance gains, with AP improvements of 3.7\% (FSVOD-500), 5.3\% (FSYTV-40), 4.3\% (VidOR), and 4.5\% (VidVRD) in the 5-shot setting. Further results demonstrate improvements in 1-shot, 3-shot, and 10-shot configurations. We make the code public at: \url{https://github.com/yogesh-iitj/fs-video-vit}.
\end{abstract}

\section{Introduction}
\begin{figure}[!t]
    \centering
      \includegraphics[width=0.47\textwidth]{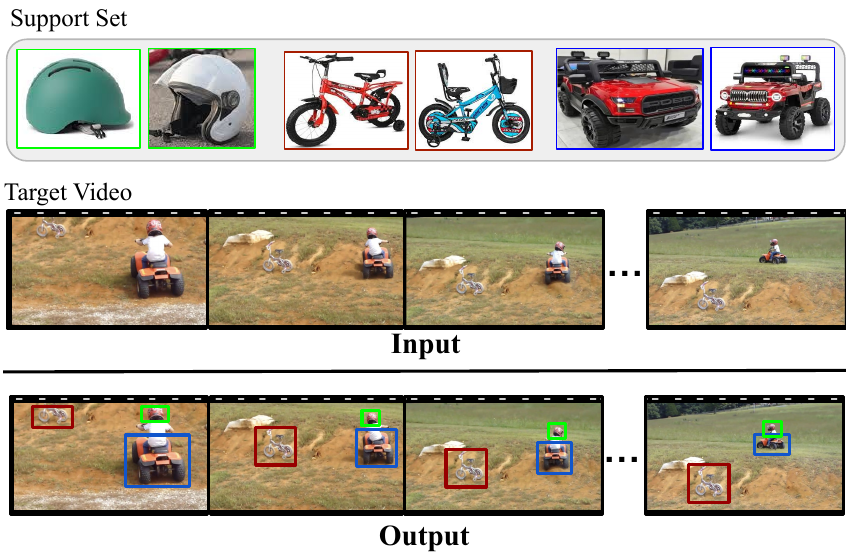}
    \caption{\label{fig:goal} Given a support set containing novel objects and a target video, the goal of \textsc{FSVOD}~\cite{fsvod} is to detect all the instances of novel objects on the target video. We propose an object-aware temporally consistent few-shot object detection framework that significantly improves the state of the art for this task.}
 
% https://docs.google.com/drawings/d/1v0SyfkqIKe5ZrsHtFojH_GakNtRKTeOs0DkVhc6WFPA/edit
\end{figure}

Object detection in videos has witnessed significant progress with the advent of deep learning-based methods in the last few years~\cite{zhu2017flow,wu2019sequence,yu2021few,fan2022few}. However, most traditional methods focus on closed-set detection, where models are trained on predefined object categories with extensive labeled samples per category. This makes them impractical for many real-world applications that require the detection of novel object categories with limited training samples. Few-Shot Video Object Detection (\textsc{fsvod})~\cite{fan2022few} addresses this challenge by enabling models to detect novel objects in videos using only a few examples, as shown in Figure~\ref{fig:goal}. 

Unlike widely-studied few-shot object detection in images~\cite{Dong2022IncrementalDETRIF,wang2020frustratingly,kang2019few,wu2020multi,sun2021fsce}, \textsc{fsvod} must account for temporal consistency, motion blur, occlusion, and variations in object appearance between frames, making the task substantially more difficult~\cite{fan2022few}. Existing approaches for \textsc{fsvod}~\cite{fsod,CenterTrack, han2023temporal, Kumar2024} face several limitations, such as they often process frames independently without leveraging temporal information~\cite{fsod}, struggle with false positives due to weak temporal matching~\cite{fsvod}, use region or tube proposal networks not optimized for few-shot settings~\cite{fsod,fsvod,tacf}, and have difficulties in discriminating with visually-similar objects and occlusions~\cite{Kumar2024}.

Recent advances in vision-language and video grounding models~\cite{video-owlvit23,Jung_2025_WACV,Kumar023-2,kumar2025moment,kumar2025matr} have shown strong capabilities in open-set video object detection and grounding. Video OWL-ViT~\cite{video-owlvit23}, built on CLIP’s vision-language pretraining, can be naturally adapted for few-shot video object detection by replacing text prompts with visual prompts from support images, enabling detection of novel objects from just a few examples. However, this adaptation lacks object-aware temporal modeling, as it processes frames independently of the detected objects, leading to inconsistent detection across time.

Our approach leverages a vision-language pretrained frame encoder to transfer semantic knowledge from image-text data, enabling recognition of novel object categories. Furthermore, contrastively-trained vision encoders are more effective in distinguishing between visually similar objects across different categories and handle partially visible or occluded objects compared to CNN backbones (like ResNet~\cite{resnet}) used in prior methods~\cite{fsvod, tacf, Kumar2024}. This robust feature representation serves as the foundation for our temporal consistency mechanism. To enhance this consistency, we fuse representations of previously detected objects with current frame detections, preserving relevant temporal information and adapting to appearance changes over time. Unlike methods that process frames independently, our approach maintains memory of past observations for better temporal reasoning. Our detection is directly optimized to condition on few-shot visual examples, providing better results compared to traditional region proposal networks. Further, unlike conventional tracking-by-detection pipelines~\cite{fsvod,tacf}, our end-to-end learning of video-specific object representations leads to superior generalization.

In summary, we make the following contributions: (i) We revisit the relatively under-explored task of Few-Shot Video Object Detection (\textsc{fsvod}) and introduce simple yet effective task-specific innovations. Our key contribution is a novel object-aware temporal modeling approach, that selectively propagates high-confidence object features across frames. This enables efficient feature progression, reducing noise accumulation, and improving detection accuracy in few-shot video scenarios.
(ii) For the first time, we adapt large-scale pre-trained vision-language models for \textsc{fsvod}, aligning with recent trends in open-world detection. This integration enables novel object detection by leveraging broad visual-linguistic knowledge, facilitating efficient adaptation and strong generalization to unseen categories. (iii) We perform extensive evaluations against baselines and state-of-the-art methods on four public \textsc{fsvod} benchmarks. Our approach achieves significant performance gains, with AP improvements of 3.7\% (FSVOD-500), 5.3\% (FSYTV-40), 4.3\% (VidOR), and 4.5\% (VidVRD) in the 5-shot setting. Further, we observe consistent improvements across 1-shot, 3-shot, and 10-shot setups, demonstrating robust generalization across different few-shot scenarios.

\section{Related Work}
\noindent{\textbf{Few-Shot Object Detection}}:
Recent advances in Few-Shot Object Detection (\textsc{fsod}) have been driven by attention mechanisms, transformers, and meta-learning techniques. Approaches like CoAE~\cite{hsieh2019one} and OS2D~\cite{osokin2020os2d} have pushed the boundaries of one-shot object detection in images through co-attention mechanisms and versatile one-stage approach, respectively. For traditional \textsc{fsod}, methods such as feature reweighting~\cite{kang2019few}, contrastive proposal encoding~\cite{sun2021fsce}, and hierarchical learning~\cite{she2022fast} have demonstrated improved performance across various benchmarks. In addition, several DETR variants~\cite{Dong2022IncrementalDETRIF,dai2021up} have emerged and shown promising results for few-shot object detection in images. 
Although \textsc{fsod} in images provides the foundational techniques and can be trivially extended to videos by performing few-shot detection on frames independently. 
Our experiments suggest that they often fall short. The inherent temporal nature of videos introduces additional complexities that require specialized approaches to maintain consistency across frames while preserving the few-shot learning paradigm. We introduce object-aware temporal consistency to overcome the inherent challenges specific to video.

While few-shot object detection in images has been extensively studied, its video counterpart remains relatively underexplored. Qi et al.~\cite{fsvod} have formally introduced a few-shot video object detection task and accompanying datasets. They proposed a proposal-based approach, where object detection trajectories are first proposed and later refined using the matching network. Kumar et al.~\cite{Kumar2024} have specifically studied this problem for one-shot settings using a novel query-guided variant of DETR. More recently, Wentano et al.~\cite{tacf} enhanced temporal information by context fusion. Our work addresses key limitations of existing \textsc{fsvod} methods by utilizing a transformer-based backbone (OWL-ViT) pre-trained on large-scale image-text pairs, replacing the CNN-based backbones used in prior approaches and improves semantic knowledge transfer. In addition, unlike proposal-based networks not optimized for few-shot scenarios, our proposal-free method conditions detection directly on few-shot-trained heads and improves generalization to novel object categories.

\noindent{\textbf{Open-World Object Detection}}: There is a fundamental shift in computer vision literature toward detecting novel visual concepts driven by models pre-trained on large-scale vision and language data~\cite{llm_fsod,mml_clas}. These large models help in detecting novel objects by leveraging their broad visual and linguistic knowledge, efficient adaptation, and improved generalization to unseen categories. Recent works like OWL-ViT~\cite{minderer2022simple} and Video OWL-ViT~\cite{heigold2023video} leverage vision-language models pretrained on large-scale data to detect open-vocabulary concepts. Methods like GLIP~\cite{li2022grounded} and OpenSeed~\cite{zhang2023simple} further advance this paradigm by incorporating grounded pretraining and structured knowledge distillation. 

This growing trend highlights the importance of flexible, prompt-driven detection systems that can adapt to diverse visual inputs. Open-world object detectors can be extended for image-conditioned detection, enabling users to detect objects by providing a few visual examples. In this work, we leverage OWL-ViT encoders for image-conditioned detection in a few-shot setup and enhance the existing Video OWL-ViT~\cite{video-owlvit23} by introducing an object-aware decoder that significantly improves temporal consistency across video frames, and thereby the few-shot video object detection performance.

\section{Method}
Few-shot video object detection (\textsc{fsvod}) aims to detect novel object categories that were not encountered during training. The task involves a \textit{support set} containing a few images of novel object classes and a \textit{target video} consisting of multiple frames. The goal is to identify and localize all instances of the objects from the support set across every frame of the target video, as shown in Figure~\ref{fig:goal}. 

\subsection{Problem Formulation}
In the \textsc{fsvod} task, we aim to locate and classify all instances of objects belonging to novel classes in a target video using only a few support examples. Formally, given an $N$-way $K$-shot support set $\mathcal{S} = \{(\mathbf{I}_{i,j}, c_i) | i = 1,\cdot,N; j = 1,\cdot,K\}$ where $\mathbf{I}_{i,j}$ represents the $j$-th support image of class $c_i$, and a target video $\mathcal{V} = \{\mathbf{F}_t | t = 1,...,T\}$ with $T$ frames, our goal is to detect all instances of the $N$ novel classes across all frames. For each frame $\mathbf{F}_t$, the model predicts a set of object bounding boxes along with their class labels.

\subsection{Design Motivation and Architectural Overview}
Our proposed approach is designed to address the following unique challenges associated with \textsc{fsvod} task: (i) the limited supervision requiring efficient knowledge transfer, (ii) the necessity for temporal consistency across sequential frames despite appearance variations due to motion, occlusion, and viewpoint shifts, and (iii) the open-set recognition capability to distinguish between novel objects and background elements without explicit training.

\begin{figure}[t!]
    \centering
      \includegraphics[width=\linewidth]{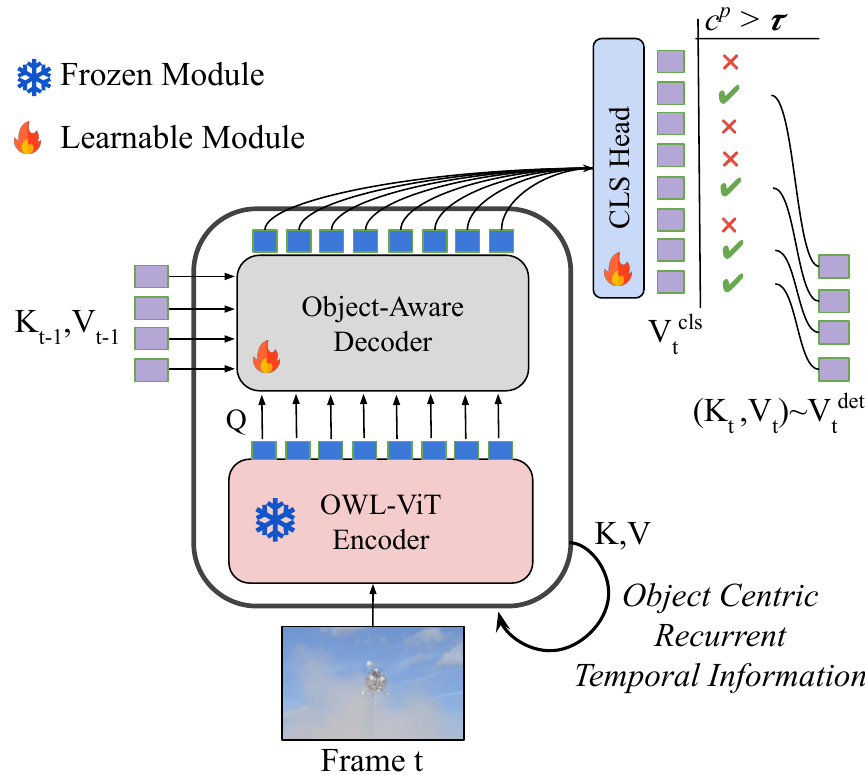}
  \caption{\label{fig:model1}Object-aware Temporal Consistency. We utilize the OWL-ViT encoder output tokens as queries but, critically, only forward the matched object tokens ($c^p > \tau$) as key-value pairs to the next frame's decoder. This mechanism selectively propagates high-confidence object features across frames. This enables efficient feature progression, reducing noise accumulation. We empirically found a threshold ($\tau$) and compared it with class probability ($c^p$) scores for selective propagation. This selective propagation mechanism significantly reduces noise accumulation across frames and maintains focused, consistent visual representations of detected objects throughout the video sequence.}
\end{figure}

\begin{figure}[!t]
    \centering
    \includegraphics[width=\linewidth]{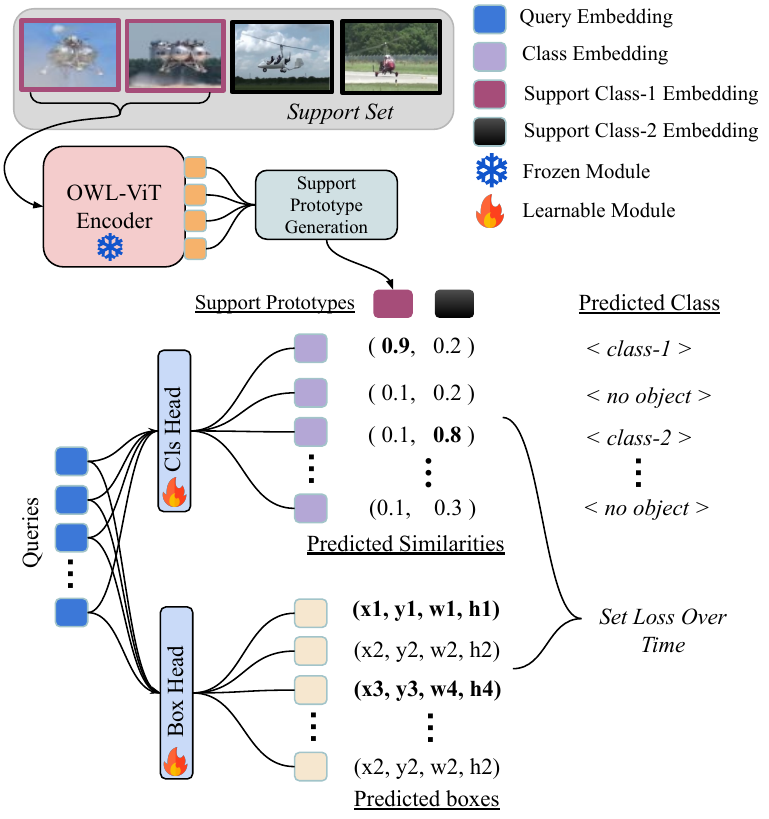}
    \caption{\label{fig:model2}Few-shot Classification and Detection Heads. Our architecture processes object queries through parallel projection heads for classification and localization. Classification embeddings are compared against support embeddings via cosine similarity, while the detection head predicts bounding box coordinates. The temporal consistency is maintained by propagating matched object queries across frames.}
\end{figure}

To address these challenges, we propose an object-aware, temporally consistent few-shot video object detection framework. Our approach consists of the following main components: a language-aligned vision encoder that provides semantically rich visual representations, a temporal fusion decoder that selectively propagates high-confidence object features across frames and, few-shot matching and detections heads that aligns target frame features with support examples. Figures~\ref{fig:model1},\ref{fig:model2} illustrates our overall framework.

\subsection{Object-aware Temporally consistent FSVOD framework}
\subsubsection{Language-Aligned Vision Encoder:}
\label{sec:l_a_enc}
Traditional CNN backbones like ResNet~\cite{resnet}, trained solely on visual classification tasks, often struggle to generalize to novel classes with limited examples. To overcome this limitation, the recent trend~\cite{llm_fsod,video-owlvit23} is to use a pretrained vision encoder that has been aligned with language semantics through large-scale vision-language pretraining. %This language-aligned encoder offers stronger semantic richness, capturing relationships between visual concepts and enabling generalization to unseen classes based on visual similarity. It demonstrates improved discriminative power, distinguishing between visually similar objects from different categories while grouping visually diverse instances of the same class.
We adopt this by utilizing an OWL-ViT encoder pretrained on large-scale image-text pairs. This encoder transforms each input image $\mathbf{I}$ into a grid of patch embeddings $\mathbf{E} \in \mathbb{R}^{P \times D}$, where $P$ is the number of patches and $D$ is the embedding dimension.

\subsubsection{Support Set Encoding:}

\textbf{(i) Support Feature Extraction.} 
Given a support set $\mathcal{S}$ containing $K$ examples for each of the $N$ novel classes, we extract patch-level features from each support image. For each support image $\mathbf{I}_{i,j}$, the encoder generates both patch embeddings, $\mathbf{E}_{i,j} \in \mathbb{R}^{P \times D},$ and their corresponding objectness scores for each patch,  $\mathbf{s}_{i,j} \in \mathbb{R}^P$. 
% \begin{equation}
% (\mathbf{E}_{i,j}, \mathbf{s}_{i,j}) = \text{Encoder}(\mathbf{I}_{i,j}), \quad \mathbf{E}_{i,j} \in \mathbb{R}^{P \times D}, \mathbf{s}_{i,j} \in \mathbb{R}^P.
% \end{equation}
A higher objectness score indicates a higher probability of an object being present in that particular patch. To obtain an object-centric representation, we select the patch with the highest objectness score.
\begin{equation}
p^* = \arg\max_p s_{i,j}^p, \quad \mathbf{z}_{i,j} = \mathbf{e}_{i,j}^{p^*},
\end{equation}
where $\mathbf{z}_{i,j}$ is the selected patch embedding that best represents the target object in the support image.
\textbf{(ii) Support Prototype Generation.} After extracting object-centric features ($\mathbf{z}_{i,j}$) from each support image, we aggregate information from the $K$ support examples to form a class prototype. For each class $c_i$, the final class prototype $\mathbf{z}_i$ is computed by averaging across the $K$ selected object-centric features.
% \begin{equation}
% \mathbf{z}_i = \frac{1}{K}\sum_{j=1}^{K} \mathbf{z}_{i,j}
% \end{equation}
\noindent This aggregation creates a robust representation for each novel class that captures essential visual characteristics across multiple support examples, allowing the model to handle intra-class variations more effectively.

\subsubsection{Object-Aware Temporal Frame Decoder:}
\label{sec:o_temp_mod}
\textbf{(i) Target Frame Feature Extraction.} For each frame $\mathbf{F}_t$ in the target video, we extract patch embeddings using the OWL-ViT image encoder, $\mathbf{F}_t \in \mathbb{R}^{P \times D}$.
% \begin{equation}
% \mathbf{F}_t = \text{Encoder}(\mathbf{F}_t) \in \mathbb{R}^{P \times D}
% \end{equation}
\textbf{(ii) Temporal Fusion.} To ensure temporal consistency in object detection across frames, we propose a filtering mechanism that selectively propagates high-confidence object features from one frame to the next. This approach facilitates efficient feature progression and minimizes noise accumulation by leveraging the embeddings of successfully detected objects from previous frames. By doing so, it enhances temporal consistency and improves the overall accuracy of object detection in video sequences.

For the first frame ($t=0$), we process the frame embeddings directly. For subsequent frames ($t>0$), we incorporate information from previous frame detections via a cross-frame interaction that selectively integrates object-level features to enhance temporal consistency. Specifically, we select embeddings from the few-shot classification head ($\mathbf{V}_t^{cls}$) of the previous frame where detection confidence exceeds a threshold $\tau$ (shown in Figure~\ref{fig:model1}):
\begin{equation}
\mathbf{V}_{t-1}^{\text{det}} = \{\mathbf{v}_{t-1}^p |\hat{c}_{t-1,i}^p > \tau\},
\end{equation}
where $\hat{c}_{t-1,i}^p$ represents the classification probability for patch $p$ and class $i$ in frame $t-1$. The current frame embeddings $\mathbf{F}_t$ serve as queries, while the embeddings from Cls Head with high confidence score responsible for object detection (Figure ~\ref{fig:model1}), $\mathbf{V}_{t-1}^{\text{det}}$ from the previous frame serve as both keys and values. The cross-frame interaction operation is computed as:
\begin{equation}
\mathbf{A}_t = \text{Softmax}\left(\frac{\mathbf{F}_t (\mathbf{V}_{t-1}^{\text{det}})^T}{\sqrt{D}}\right).
\end{equation}

\noindent The resulting attention-weighted features are used to update the current frame representations:
\begin{equation}
\hat{\mathbf{F}}_t = \mathbf{F}_t + \mathbf{A}_t\mathbf{V}_{t-1}^{\text{det}}.
\end{equation}

\noindent These temporally enhanced features $\hat{\mathbf{F}}_t$ are then fed to the classification and localization heads for object detection in the current frame. By using attention between current frame features and previously detected objects representations, the model maintains better temporal consistency while detecting objects in the current frame.

\subsubsection{Few-Shot Detection Heads:}
\label{sec:fs_heads}
\noindent The temporally enhanced frame embeddings $\hat{\mathbf{F}}_t$ are processed through parallel classification and localization heads to produce the final detection results.

\noindent \textbf{Classification Head.} The classification head compares frame embeddings with support class prototypes to determine object categories. It projects frame embeddings to the classification space:
\begin{equation}
\mathbf{V}_t^{cls} = \mathbf{W}_{\text{cls}}\hat{\mathbf{F}}_t,
\end{equation}
where $\mathbf{W}_{\text{cls}}$ is a learnable projection matrix.

\noindent For each patch embedding $\mathbf{v}_t^p$ and class prototype $\mathbf{z}_i$, we compute cosine similarity:
\begin{equation}
s_{p,i} = \frac{\mathbf{v}_t^p \cdot \mathbf{z}_i}{||\mathbf{v}_t^p|| \cdot ||\mathbf{z}_i||}.
\end{equation}

\noindent Each patch is assigned a probability distribution over the $N$ classes and background using softmax normalization of similarity scores, translating feature similarity into class probabilities.

\noindent \textbf{Localization Head.} The localization head predicts bounding box coordinates for detected objects through a multi-layer perceptron:
\begin{equation}
\hat{\mathbf{B}}_t = \text{MLP}_{\text{box}}(\hat{\mathbf{F}}_t),
\end{equation}
where each row $\hat{\mathbf{b}}_t^p \in \mathbb{R}^4$ represents the predicted bounding box parameters $(x, y, w, h)$ for each corresponding patch $p$. For each patch $p$, if its maximum classification score exceeds threshold $\kappa$, we consider it as a valid detection.

\subsubsection{Training Objective:}
\noindent Our model is trained end-to-end by optimizing a combined loss function across all video frames:

$$
\begin{aligned}
\mathcal{L} = \sum_{t=1}^{T} \Big[ & \sum_{m=1}^{M_t} \lambda_{\text{cls}} \cdot \mathcal{L}_{\text{cls}}(\hat{\mathbf{c}}_t^{\sigma_t(m)}, c_t^m) \\
& + \sum_{m=1}^{M_t} \lambda_{\text{box}} \cdot \mathcal{L}_{\text{box}}(\hat{\mathbf{b}}_t^{\sigma_t(m)}, \mathbf{b}_t^m) \Big].
\end{aligned}
$$

\noindent The classification loss (\(\mathcal{L}_{\text{cls}}\)) uses cross-entropy to measure discrepancy between predicted and ground-truth class labels, while the localization loss (\(\mathcal{L}_{\text{box}}\)) combines L1 loss for coordinate distances and generalized IoU loss for handling varying box sizes. Following DETR~\cite{detr}, $\sigma_t(m)$ denotes the index of the prediction matched to ground-truth object $m$, and $M_t$ is the number of objects in frame $t$. The hyperparameters $\lambda_{\text{cls}}$ and $\lambda_{\text{box}}$ control the weighting of the classification and localization losses.

\begin{table*}[!th]
\scriptsize
\centering
\begin{tabular}{ll ccc ccc ccc ccc}
\toprule
&\multirow{2}{*}{Method}  & \multicolumn{3}{c}{FSVOD-500} & \multicolumn{3}{c}{FSYTV-40} & \multicolumn{3}{c}{VidOR} & \multicolumn{3}{c}{VidVRD} \\
\cmidrule(lr){3-5} \cmidrule(lr){6-8} \cmidrule(lr){9-11} \cmidrule(lr){12-14}
& & $AP$ & $AP_{50}$ & $AP_{75}$ & $AP$ & $AP_{50}$ & $AP_{75}$ & $AP$ & $AP_{50}$ & $AP_{75}$ & $AP$ & $AP_{50}$ & $AP_{75}$\\
\midrule
\multirow{7}{*}{\rotatebox{90}{\textbf{Non-temporal }}}&
FR-CNN  & 18.2 & 26.4 & 19.6 & 9.3 & 15.4 & 9.6  & 21.6 & 29.1 & 23.2 & 14.3 & 18.6 & 14.2 \\
&FSOD & 21.1 & 31.3 & 22.6 & 12.5 & 20.9 & 13.0 & - & - & - & - & - & - \\
&RDN & 18.2 & 27.9 & 19.7 & 8.1 & 13.4 & 8.6 & - & - & - & - & - & - \\
&CoAE &18.4&29.3&18.9&9.8&18.4&10.1&35.8&43.7&36.1&30.2&38.1&31.8\\
&Retrieval-Based &1.3&2.4&1.6&0.6&1.1&0.8&1.4&2.6&1.3&1.8&2.5&2.0\\
&Owl-ViT &14.8&25.3&16.7&10.6&19.8&11.2&30.1&37.3&31.2&23.6&32.1&23.9\\
&UP-DETR &20.1&27.6&21.2&11.8&19.4&12.1&39.7&47.5&40.0&11.6&18.3&12.5\\
\midrule
\multirow{12}{*}{\rotatebox{90}{\textbf{Temporal}}}
&DeepSort+YOLOv8 + CLIP &13.2&21.4&14.4&5.3&9.4&6.2&23.2&30.1&23.6&23.4&33.1&24.1\\
&TrackFormer + CLIP &14.2&22.6&15.3&6.2&10.5&6.8&25.3&32.7&26.2&25.4&33.8&27.3\\
&ByteTrack  + CLIP &14.7&23.9&15.8&7.9&15.6&8.1&24.9&31.4&25.3&25.6&36.2&26.2\\
&MEGA  & 16.8 & 26.4 & 17.7 & 7.8 & 13.0 & 8.3 & - & - & - & - & - & -  \\
&CTracker  & 20.1 & 30.6 & 21.0 & 8.9 & 14.4 & 9.1&-&-&-&-&-&- \\
&FairMOT & 20.3 & 31.0 & 21.2 & 9.6 & 16.0 & 9.5&-&-&-&-&-&- \\
&CenterTrack & 20.6 & 30.5 & 21.9 & 9.5 & 15.6 & 9.7&-&-&-&-&-&- \\
&TACF& \underline{26.9} & \underline{39.2} & \underline{27.9} & \underline{15.9} & 24.2 & \underline{17.9} &-&-&-&-&-&-\\
&Video Owl-ViT  & 25.8 & 36.5 & 26.1 & 15.7 & \underline{26.3} & 15.9 & 44.3 & 52.9 & 45.7 & \underline{44.2} & \underline{53.7} & \underline{45.4}\\
&FSVOD &  25.1 & 36.8 & 26.2 & 14.6 & 21.9 & 16.1& \underline{45.1}& \underline{54.3}& \underline{46.2}&40.7&53.5&43.7\\
&QDETRv & 26.1 & 33.8 & 23.7 & 14.1 & 22.8 & 15.3 &43.4&51.8&44.6& 42.8& 50.8& 41.7 \\
&Ours & \textbf{30.6} & \textbf{42.9} & \textbf{32.1} & \textbf{21.2} & \textbf{29.8} & \textbf{23.5} &\textbf{49.4}&\textbf{57.3}&\textbf{50.2}&\textbf{48.7}&\textbf{57.8}&\textbf{49.2} \\
\bottomrule
\end{tabular}
\caption{Performance comparison on FSVOD-500, FSYTV-40, VidOR, and VidVRd in 5-shot set-up. Our Object-aware Video OWL-ViT consistently surpasses both state-of-the-art and baseline methods across all datasets and metrics.}

\label{tab:results}
\end{table*}

\begin{table}[t!]
\scriptsize
\centering
\begin{tabular}{lcccc}
\toprule
Split & Vid-OR & VidVRD & FSVOD & FSYTV\\
\midrule
\#Train Videos & 6,164 & 758 & 2553 & 1627 \\
\#Train Object Queries & 57,599 & 4,395 & - & - \\
\#Train Object Categories & 75 & 30 & 320 & 30 \\
\#Test Videos & 32 & 42 & 949 & 608 \\
\#Test Object Queries & 407 & 112 & - & - \\
\#Test Object Categories & 9 & 5 & 100 & 10\\
\bottomrule
\end{tabular}
\caption{Overview of the datasets used in this work.}
\label{tab:dataset_summary}
\end{table}

\section{Experiments and Results}

\subsection{\textbf{Datasets}}
\label{sec:dataset} 
In this work, we utilized the FSVOD-500, FSYTV-40, VidOR, and VidVRD datasets, as proposed and repurposed in~\cite{fsvod,Kumar2024}. These datasets are specifically designed for few-shot video object detection, providing a structured evaluation framework with disjoint training, validation, and testing classes. Detailed statistics of the datasets are provided in Table~\ref{tab:dataset_summary}.

\subsection{Baselines}

We compare our method against temporal and non-temporal few-shot object detectors, covering diverse architectural designs and learning strategies.

\noindent \textbf{Non-temporal Methods.} We establish strong baselines using image-based few-shot detectors compared in~\cite{fsvod}. Traditional detectors like Faster R-CNN~\cite{FasterRCNN} validate core detection performance, while FSOD~\cite{fsod} and RDN~\cite{RDN} from~\cite{fsvod} highlight the impact of specialized few-shot designs. We adapt CoAE~\cite{hsieh2019one}, originally for one-shot detection, by selecting high-confidence predictions for few-shot extension. CoAE employs meta-learning and attention for data efficiency. To assess large-scale pre-training, we include OWL-ViT~\cite{owl} and UP-DETR~\cite{wang2021unsupervised}. Retrieval-based detection~\cite{radenović2016cnn} is also evaluated as a proposal selection strategy.

\noindent \textbf{Temporal Methods.} We evaluate video-based detectors across three categories. First, we adopt temporal baselines from~\cite{fsvod}, including MEGA~\cite{MEGA}, CTracker~\cite{CTracker}, FairMOT~\cite{FairMOT}, and CenterTrack~\cite{CenterTrack}. Second, we assess tracking-by-detection methods, DeepSort+YOLOv8~\cite{deepsort}, TrackFormer~\cite{trackformer}, and ByteTrack~\cite{bytetrack} combined with CLIP~\cite{clip} for tube-support similarity. Models are trained on support examples to generate spatio-temporal tubes, and tube-support similarity is computed via average pooling. Third, we compare with SoTA one/few-shot video detectors, including FSVOD~\cite{fsvod}, TACF~\cite{tacf}, and QDETRv~\cite{Kumar2024}. Lastly, we adapt Video OWL-ViT~\cite{video-owlvit23} by replacing its text-encoder with OWL-ViT’s image-encoder, integrating our few-shot detection and classification heads.

\begin{table*}[!th]
\scriptsize
\centering
 \begin{tabular}{llcccccccccccc}
\toprule
&\multirow{2}{*}{Method}& \multicolumn{3}{c}{FSVOD-500} & \multicolumn{3}{c}{FSYTV-40} & \multicolumn{3}{c}{VidOR} & \multicolumn{3}{c}{VidVRD}\\
\cmidrule(lr){3-5} \cmidrule(lr){6-8} \cmidrule(lr){9-11} \cmidrule(lr){12-14}
       && $AP$ & $AP_{50}$ & $AP_{75}$ & $AP$ & $AP_{50}$ & $AP_{75}$ & $AP$ & $AP_{50}$ & $AP_{75}$ & $AP$ & $AP_{50}$ & $AP_{75}$  \\
\midrule
%  \rowcolor{gray!15} \multicolumn{13}{c}{{\textbf{1-Shot}}}\\
% \midrule
\multirow{7}{*}{\rotatebox{90}{\textbf{1-Shot}}}
&CoAE &17.2&27.8&17.2&6.7&17.3&9.8&33.2&42.5&35.2&29.6&37.6&31.5\\
&UP-DETR &17.3&24.3&18.4&9.7&16.9&9.9&35.6&45.2&37.2&8.5&15.2&9.3\\
&DeepSort+YOLOv8 + CLIP &12.2&20.1&12.8&4.1&7.9&4.7&21.4&28.4&22.7&21.9&31.7&22.8\\
&FSVOD    &20.7 &29.4 &21.4 &11.4 & 18.6 & 12.1& 36.8 & 45.3&36.3 &34.5 &45.2  &35.2       \\
&\model   & \underline{25.2} & \underline{34.9}  & \underline{25.8} & \underline{13.6} & \underline{24.1}  & \underline{14.2} & \underline{41.8} & 50.3 & \underline{42.2} & \underline{40.3} & \underline{49.7} & 40.1   \\
&Video OWL-ViT & 23.8 & 33.8 & 24.7 & 12.4 & 22.7& 13.2& 40.3& \underline{51.7} & 41.5 & 39.8 & 47.6& \underline{40.2}\\
&Ours &\textbf{27.4} & \textbf{39.1}& \textbf{28.9}& \textbf{18.3}& \textbf{26.2}&\textbf{19.7} &\textbf{45.2} &\textbf{54.1} &\textbf{47.6} &\textbf{45.3} &\textbf{54.5} & \textbf{46.8} \\
\midrule
\multirow{7}{*}{\rotatebox{90}{\textbf{3-Shot}}}
&CoAE &17.8&28.2&17.6&6.9&17.5&9.8&33.3&42.6&35.6&29.9&37.8&31.9\\
&UP-DETR &17.9&25.1&18.8&10.2&17.5&10.3&36.2&46.8&38.3&9.3&16.1&10.4\\
&DeepSort+YOLOv8 + CLIP  &12.7&20.4&13.7&4.3&8.3&4.9&22.0&28.8&23.1&22.0&31.9&23.4\\
&FSVOD   &21.9 &30.8 &22.1 &12.3 & 19.2 & 12.9& 37.6 & 46.6 &37.8 &35.4 &46.1  &36.4       \\
&\model  & \underline{25.6} & \underline{35.2}  & \underline{26.3}  & \underline{13.8} & \underline{24.5}  & \underline{14.5} & \underline{42.1} & 50.8 & \underline{42.7} & \underline{40.8} & \underline{50.3} & 40.6  \\
& Video OWL-ViT & 24.1 & 34.6& 25.2 & 12.7 & 23.2  & 13.8 & 40.6& \underline{52.5} &41.8 &40.4 & 48.3 & \underline{40.7}    \\
& Ours &\textbf{28.2} & \textbf{40.5}& \textbf{31.1}& \textbf{19.6}& \textbf{27.2}&\textbf{20.9} &\textbf{46.4} &\textbf{55.6} &\textbf{48.2} &\textbf{46.5} &\textbf{55.7} & \textbf{47.2} \\
\midrule
\multirow{7}{*}{\rotatebox{90}{\textbf{10-Shot}}}
&CoAE &20.1&29.9&19.7&10.2&19.8&12.3&37.2&44.9&38.2&32.3&39.7&32.7\\
&UP-DETR &23.2&30.2&23.7&13.7&22.3&15.2&41.8&50.2&43.1&13.7&21.2&15.4\\
&DeepSort+YOLOv8 + CLIP &16.6&23.7&17.2&7.8&11.5&9.4&26.3&32.9&25.8&25.1&35.6&26.6\\
&FSVOD &  \underline{27.2} & \underline{38.5} & \underline{28.6} & \underline{16.1} & \underline{25.8} & \underline{17.9}& \underline{47.3}& \underline{56.7}& \underline{48.4}& \underline{44.7}& \underline{54.9}& \underline{44.9}\\
&\model{} & 26.2 & 33.8 & 23.8 & 14.2 & 23.1 & 15.6 &43.9&52.1&45.2&43.1&53.8&44.2   \\
&Video OWL-ViT & 26.7 & 35.8 & 27.1 & 14.2 & 25.1& 14.9 & 43.7 & 54.2& 44.1 & 43.7 & 52.8 & 44.3   \\
& Ours & \textbf{33.2} & \textbf{45.2} & \textbf{34.9} & \textbf{23.5} & \textbf{32.2} & \textbf{25.2} &\textbf{51.8}&\textbf{59.6}&\textbf{53.3}&\textbf{50.3}&\textbf{59.2}&\textbf{51.7}  \\ 
\bottomrule
\end{tabular}
\caption{Few-shot performance comparison. Our method consistently outperforms SoTA and implemented baselines across all settings and datasets, demonstrating its effectiveness across diverse settings and datasets.}
\label{tab:few-shot}
\end{table*}

\begin{figure}[!th]
    \centering
    \includegraphics[width=0.8\linewidth]{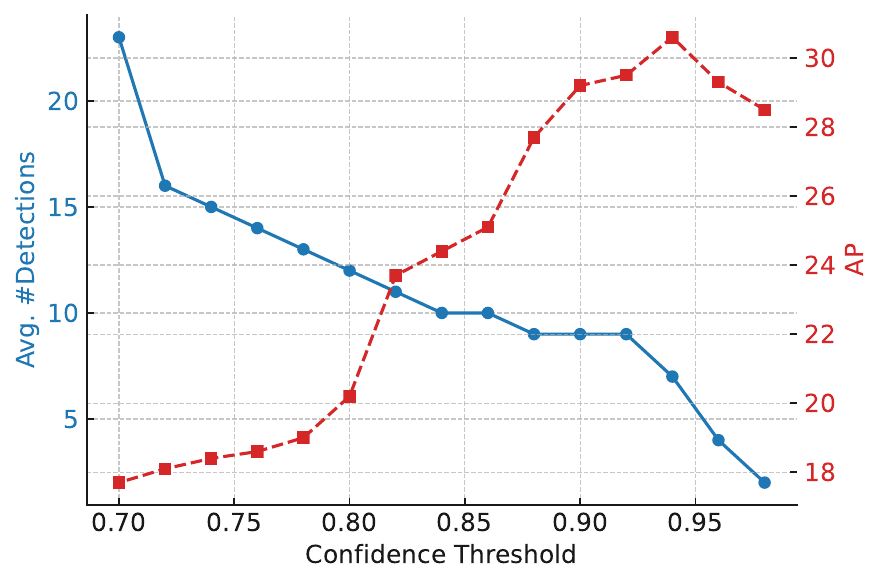}
    \caption{\label{fig:thr_oadec} Threshold effect ($c^p > \tau$) during the object-aware temporal consistency. }
\end{figure}

\subsection{\textbf{Performance Measures}} 
Following~\cite{fsvod}, we adopt Average Precision (AP) as an evaluation metric, computed as the area under the precision-recall curve. AP$_{50}$ and AP$_{75}$ denote performance at IoU thresholds of 50\% and 75\%, respectively.

\subsection{Implementation Details}
We implement our model using pretrained OWL ViT-L/16 for the vision encoder. The temporal fusion module employs a 4-head cross-attention mechanism with 1024-dimensional hidden states. The classification and localization heads use 2-layer MLPs with 512-dimensional hidden states. During training, we use AdamW optimization with a 1e-5 learning rate, 0.01 weight decay, and cosine scheduling with a linear warmup. The loss weighting parameters are set to $\lambda_{\text{cls}}=2$, $\lambda_{\text{box}}=5$. We set $\tau = 0.94$ and $\kappa = 0.98$ in our experiments.

\subsection{Results and Discussion}
\noindent{\textbf{Main Results.}}
Our proposed approach achieves state-of-the-art performance across all four benchmark datasets. On FSVOD-500, we achieve AP of 30.6\%, AP${50}$ of 42.9\%, and AP${75}$ of 32.1\%, surpassing TACF by margins of 3.7\%, 3.7\%, and 4.2\%, highlighting robustness in localization and classification. Similarly, on FSYTV-40, our method yields 21.2\% AP, 29.8\% AP${50}$, and 23.5\% AP${75}$, outperforming TACF by 5.3\%, 5.6\%, and 5.6\%. Performance gains are more pronounced on VidOR and VidVRD, 49.4\% AP, 57.3\% AP${50}$, and 50.2\% AP${75}$ for VidOR, and 48.7\% AP, 57.8\% AP${50}$, and 49.2\% AP${75}$ for VidVRD, setting new benchmarks. These results underscore the effectiveness of our object-aware temporal modelling in capturing complex spatiotemporal dependencies with minimal supervision.

\begin{figure*}[t]
    \centering
      \includegraphics[width=0.82\textwidth]{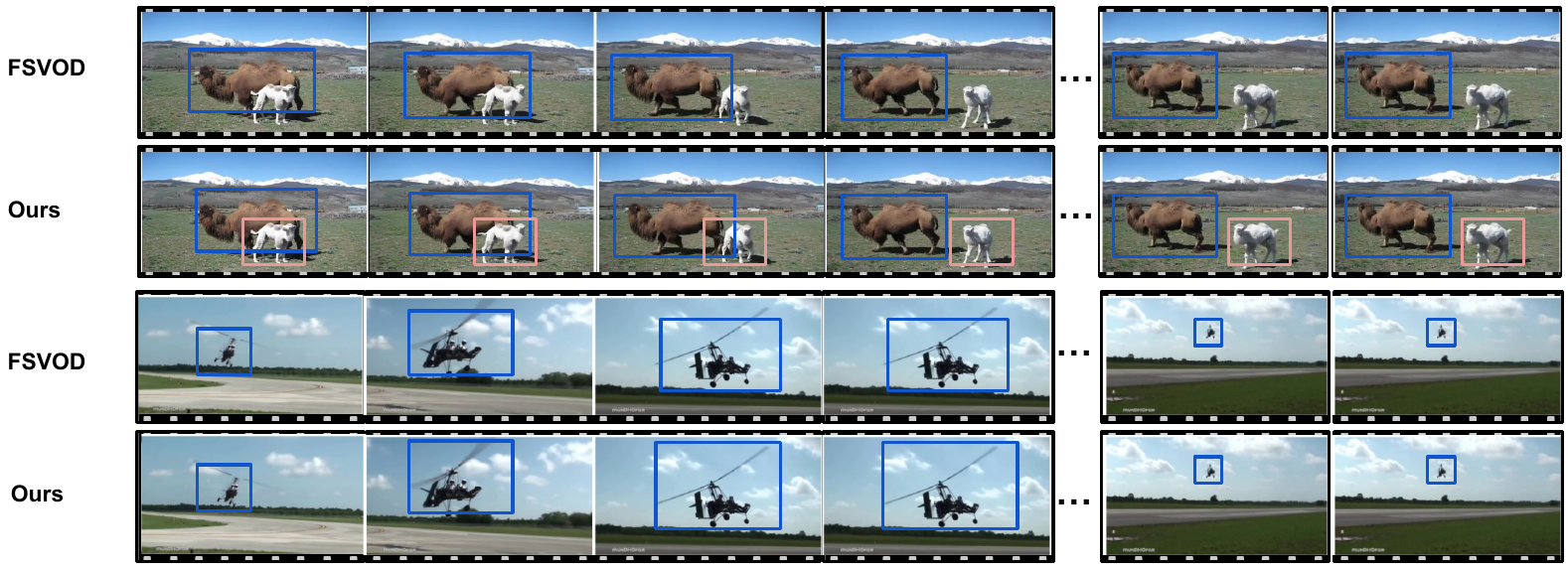}
    \caption{\label{fig:qual} Qualitative comparison FSVOD-500 test set videos. Our method accurately detects multiple visually distinct instances of \emph{Bactrian} camels (top) and precisely localizes the \emph{Autogyro} (bottom), whereas FSVOD misses several camel instances.}

  % \caption{\label{fig:qual}. Qualitative comparison of detection results with FSVOD for two test videos from FSVOD-500. Our approach accurately detects multiple instances of the \emph{Bactrian} camel (top) having differnce visual appearance and precisely localizes the \emph{Autogyro} (bottom). Notably, FSVOD miss some camel instances. Refer to supplementary for additional comparative results with other baselines.}
  % https://docs.google.com/drawings/d/1Gon_l5zg_q28MSPgb4qUjdxtE7MRlazgb-RG459edXU/edit
\end{figure*}

\begin{figure}[!th]
    \centering
    \includegraphics[width=0.84\linewidth]{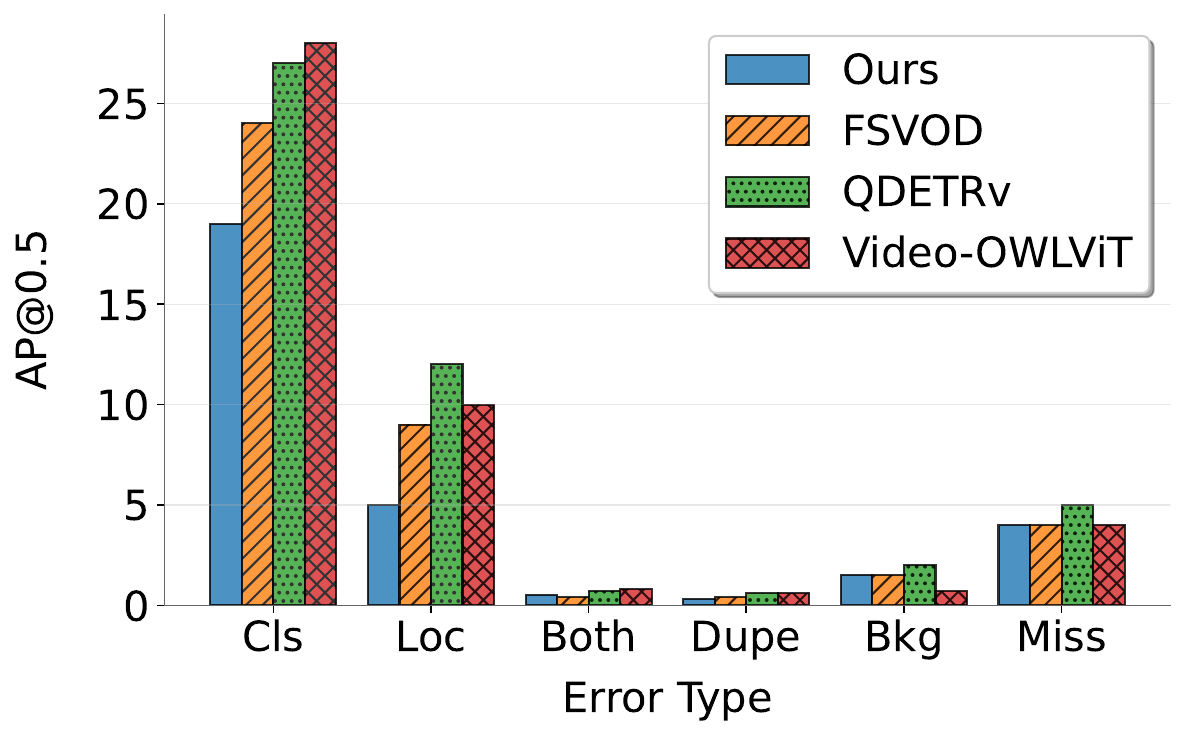}
    \caption{\label{fig:error_type} Error type analysis on FSVOD val set. A lower AP score is better. }
\end{figure}

\noindent{\textbf{Temporal vs. Non-temporal Methods.}} Temporal modeling yields consistent gains across datasets. Non-temporal methods like UP-DETR (20.1\% AP on FSVOD-500) underperform compared to our approach (30.6\%), reflecting a 10.5\% gap. On FSYTV-40, our method improves by 9.4\%. These results underscore the value of capturing temporal dependencies, especially under few-shot constraints.

\noindent{\textbf{Comparison with adapted Temporal Methods.}} Despite using strong trackers (ByteTrack, DeepSort, TrackFormer) paired with CLIP’s semantic features, these approaches fall short of specialized video detectors. ByteTrack+CLIP reaches only 14.7\% AP on FSVOD-500, underperforming our method by 15.9\%. Consistent gaps are observed across FSYTV-40 (13.3\%), VidOR (24.5\%), and VidVRD (23.1\%), highlighting the need for dedicated few-shot modules.

\noindent{\textbf{Improvements over Specialized Few-Shot Methods.}} Our approach delivers consistent gains over prior specialized few-shot video detectors, including FSVOD, TACF, and adapted Video OWL-ViT. Compared to FSVOD, we improve by 5.5\% AP on FSVOD-500, 6.6\% on FSYTV-40, 4.3\% on VidOR, and 8.0\% on VidVRD. Similar improvements are observed over TACF and Video OWL-ViT. These results highlight our advantages in temporal aggregation via the object-aware frame encoder and superior support-query matching through few-shot tailored heads.

\noindent{\textbf{Advantage of Temporal Fusion (Object-Aware Decoder).}} In Table~\ref{tab:obj-aware_decoder} on FSVOD-500, AP rises from 25.4\% to 30.6\% (+5.2\%), AP$_{50}$ from 37.1\% to 42.9\% (+5.8\%), and AP$_{75}$ from 26.9\% to 32.1\% (5.2\%) when temporal fusion used. On FSYTV-40, AP improves from 19.2\% to 21.2\% (+2.0\%), AP$_{50}$ from 26.8\% to 29.8\% (+3.0\%), and AP$_{75}$ from 21.4\% to 23.5\% (+2.1\%). These gains highlight the decoder’s enhanced temporal reasoning detection performance.

\noindent{\textbf{Effect of Threshold ($\tau$) on Temporal Propagation.}} Figure~\ref{fig:thr_oadec} illustrates how varying the confidence threshold $\tau$ impacts detection. As $\tau$ decreases from 0.98 to 0.70, the average number of detections rises from 2 to 23, reflecting increased object coverage. AP peaks at 30.6 when $\tau = 0.94$, then drops to 17.7 at $\tau = 0.70$, indicating precision loss due to false positives. Thus, $\tau = 0.94$ offers the best trade-off between coverage and precision.

\noindent{\textbf{Few-shot Performance Comparison.}} Table~\ref{tab:few-shot} shows consistent improvement of our method across baselines in 1-shot, 3-shot, and 10-shot settings. In the 1-shot case, we achieve 27.4\% AP on FSVOD-500 and 18.3\% on FSYTV-40, outperforming FSVOD by 6.7\% and 6.9\%, respectively. In the 10-shot setup, our method attains 33.2\% AP on FSVOD-500 and 23.5\% on FSYTV-40, with gains of 6.0\% and 7.4\% over FSVOD. Similar improvements across other benchmarks further validate our approach.

\begin{table}[!th]
\scriptsize
    \centering
\begin{tabular}{c c c c c c c}
\toprule
 \multicolumn{1}{l}{}  & \multicolumn{3}{c}{FSVOD-500}  & \multicolumn{3}{c}{FSYTV-40}\\
 \cmidrule(r){2-4} \cmidrule(r){5-7}
Temopral Fusion & $AP$ & $AP_{50}$ & $AP_{75}$ & $AP$ & $AP_{50}$ & $AP_{75}$\\
 \midrule 
 % \xmark & \xmark & 53.2 & 54.8 \\
 % \xmark &  \checkmark &  & \\
  \xmark  & 25.4 & 37.1 & 26.9 & 19.2 & 26.8& 21.4\\
    \cmark & \textbf{30.6} & \textbf{42.9} & \textbf{32.1} & \textbf{21.2} & \textbf{29.8} & \textbf{23.5} \\
 \bottomrule
\end{tabular}
\caption{Advantage of Object-Aware Temporal Fusion.\label{tab:obj-aware_decoder}}

\end{table}
\noindent{\textbf{Qualitative Results.}} Figure~\ref{fig:qual} presents qualitative comparisons on two FSVOD-500 videos. In the first, our method accurately detects multiple bactrian camels missed by FSVOD. In the second, involving an autogiro, our bounding boxes are more precise. \\

\noindent{\textbf{Error Type Analysis.}}
Following TIDE~\cite{tide-eccv2020}, we analyze error types in Fig.~\ref {fig:error_type}. Classification errors dominate all approaches (19-28 AP@0.5), with our method achieving lowest classification (19) and localization (5) errors. Video-OWLViT excels in background handling (0.7) but suffers from high classification errors (28), while QDETRv performs the poorest overall. Improving feature discrimination and localization precision are key challenges. \\

\noindent{\textbf{Efficiency Analysis.}}
Our method achieves a strong balance between accuracy and efficiency, processing at 14 FPS with 8.1 GB of memory while attaining 30.6 AP. This is 40\% faster than VideoOWL-ViT (10 FPS, 8.4 GB) and 27\% faster than QDETRv (11 FPS, 7.6 GB), while providing substantially higher accuracy (+4.8 AP and +4.5 AP, respectively). Although FSVOD operates at 16 FPS with lower memory usage (5.8 GB), our approach delivers a +5.5\% AP improvement with only a minor trade-off. Compared to other baselines, DeepSort runs at 9 FPS (7.9 GB), CenterTrack at 12 FPS (6.2 GB), and UP-DETR at 13 FPS (7.4 GB).

\section{Conclusion}  
In this work, we introduced a novel approach for Few-Shot Video Object Detection that incorporates a language-aligned vision encoder and a filtering mechanism to selectively propagate high-confidence object features across frames. This mechanism efficiently enhances feature progression, reduces noise accumulation, and improves detection accuracy in a few-shot video setting. Through extensive experiments on four datasets, our approach consistently outperforms state-of-the-art methods, achieving superior detection performance and demonstrating its effectiveness in real-world video object detection tasks.

\noindent\textbf{Acknowledgment:}
Yogesh Kumar is supported by a UGC fellowship, Govt. of India.

\bibliography{aaai2026}

@String(IJCV = {Int. J. Comput. Vis.})

@String(CVPR= {IEEE Conf. Comput. Vis. Pattern Recog.})

@String(ICCV= {Int. Conf. Comput. Vis.})

@String(ECCV= {Eur. Conf. Comput. Vis.})

@String(ICIP = {IEEE Int. Conf. Image Process.})

@String(AAAI = {AAAI})

@String(IJCV  = {IJCV})

@String(CVPR  = {CVPR})

@String(ICCV  = {ICCV})

@String(ECCV  = {ECCV})

@String(ICIP  = {ICIP})

@inproceedings{radenović2016cnn,
  title={CNN image retrieval learns from BoW: Unsupervised fine-tuning with hard examples},
  author={Radenovi{\'c}, Filip and Tolias, Giorgos and Chum, Ond{\v{r}}ej},
  booktitle={ECCV},
  year={2016},
}

@inproceedings{vidor,
    title={Annotating Objects and Relations in User-Generated Videos},
    author={Shang, Xindi and Di, Donglin and Xiao, Junbin and Cao, Yu and Yang, Xun and Chua, Tat-Seng},
    booktitle={ICMR},
    year={2019},
}

@inproceedings{vidvrd,
    author={Shang, Xindi and Ren, Tongwei and Guo, Jingfan and Zhang, Hanwang and Chua, Tat-Seng},
    title={Video Visual Relation Detection},
    booktitle={ACM MM},
    year={2017}
}

@inproceedings{wang2020frustratingly,
  title={Frustratingly Simple Few-Shot Object Detection},
  author={Wang, Xin and Huang, Thomas and Gonzalez, Joseph and Darrell, Trevor and Yu, Fisher},
  booktitle={ICML},
  year={2020},
}

@inproceedings{kang2019few,
  title={Few-shot object detection via feature reweighting},
  author={Kang, Bingyi and Liu, Zhuang and Wang, Xin and Yu, Fisher and Feng, Jiashi and Darrell, Trevor},
  booktitle={CVPR},
  year={2019}
}

@inproceedings{wu2020multi,
  title={Multi-scale positive sample refinement for few-shot object detection},
  author={Wu, Jiaxi and Liu, Songtao and Huang, Di and Wang, Yunhong},
  booktitle={ECCV},
  year={2020}
}

@article{yu2021few,
  title={When Few-Shot Learning Meets Video Object Detection},
  author={Yu, Zhongjie and Wang, Gaoang and Chen, Lin and Raschka, Sebastian and Luo, Jiebo},
  journal={arXiv preprint arXiv:2103.14724},
  year={2021}
}

@inproceedings{osokin2020os2d,
  title={Os2d: One-stage one-shot object detection by matching anchor features},
  author={Osokin, Anton and Sumin, Denis and Lomakin, Vasily},
  booktitle={ECCV},
  year={2020},
}

@inproceedings{owl,
author = {Minderer, Matthias and Gritsenko, Alexey and Stone, Austin and Neumann, Maxim and Weissenborn, Dirk and Dosovitskiy, Alexey and Mahendran, Aravindh and Arnab, Anurag and Dehghani, Mostafa and Shen, Zhuoran and Wang, Xiao and Zhai, Xiaohua and Kipf, Thomas and Houlsby, Neil},
title = {Simple Open-Vocabulary Object Detection},
year = {2022},
booktitle = {ECCV},
}

@inproceedings{she2022fast,
  title={Fast Hierarchical Learning for Few-Shot Object Detection},
  author={She, Yihang and Bhat, Goutam and Danelljan, Martin and Yu, Fisher},
  booktitle={IROS},
  year={2022},
}

@inproceedings{wang2021unsupervised,
  title={Unsupervised visual representation learning by tracking patches in video},
  author={Wang, Guangting and Zhou, Yizhou and Luo, Chong and Xie, Wenxuan and Zeng, Wenjun and Xiong, Zhiwei},
  booktitle={CVPR},
  year={2021}
}

@inproceedings{detr,
  title={End-to-end object detection with transformers},
  author={Carion, Nicolas and Massa, Francisco and Synnaeve, Gabriel and Usunier, Nicolas and Kirillov, Alexander and Zagoruyko, Sergey},
  booktitle={ECCV},
  year={2020},
}

@inproceedings{Dong2022IncrementalDETRIF,
  title={Incremental-DETR: Incremental Few-Shot Object Detection via Self-Supervised Learning},
  author={Na Dong and Yongqiang Zhang and Ming Ding and Gim Hee Lee},
  booktitle={AAAI},
  year={2022},
}

@InProceedings{Kumar2024, 
author={Kumar, Yogesh and Mallick, Saswat and Mishra, Anand and Rasipuram, Sowmya and Maitra, Anutosh and Ramnani, Roshni},
title={QDETRv: Query-Guided DETR for One-Shot Object Localization in Videos}, 
booktitle={AAAI}, 
year={2024},}

@inproceedings{fsvod,
author = {Fan, Qi and Tang, Chi-Keung and Tai, Yu-Wing},
title = {Few-Shot Video Object Detection},
year = {2022},
booktitle = {ECCV},

}

@inproceedings{deepsort,
  title={Simple online and realtime tracking with a deep association metric},
  author={Wojke, Nicolai and Bewley, Alex and Paulus, Dietrich},
  booktitle={ICIP},
  year={2017},
}

@inproceedings{trackformer,
  title={Trackformer: Multi-object tracking with transformers},
  author={Meinhardt, Tim and Kirillov, Alexander and Leal-Taixe, Laura and Feichtenhofer, Christoph},
  booktitle={CVPR},
  year={2022}
}

@inproceedings{bytetrack,
  title={Bytetrack: Multi-object tracking by associating every detection box},
  author={Zhang, Yifu and Sun, Peize and Jiang, Yi and Yu, Dongdong and Weng, Fucheng and Yuan, Zehuan and Luo, Ping and Liu, Wenyu and Wang, Xinggang},
  booktitle={ECCV},
  year={2022},
}

@inproceedings{clip,
  title={Learning transferable visual models from natural language supervision},
  author={Radford, Alec and Kim, Jong Wook and Hallacy, Chris and Ramesh, Aditya and Goh, Gabriel and Agarwal, Sandhini and Sastry, Girish and Askell, Amanda and Mishkin, Pamela and Clark, Jack and others},
  booktitle={ICML},
  year={2021},
}

@inproceedings{han2023temporal,
  title={Temporal Aggregation with Context Focusing for Few-Shot Video Object Detection},
  author={Han, Wentao and Lei, Jie and Wang, Fahong and Feng, Zunlei and Liang, Ronghua},
  booktitle={SMC},
  year={2023},
}

@inproceedings{FasterRCNN,
  author    = {Ren, Shaoqing and He, Kaiming and Girshick, Ross and Sun, Jian},
  title     = {Faster R-CNN: Towards Real-Time Object Detection with Region Proposal Networks},
  booktitle = {NeurIPS},
  year      = {2015}
}

@inproceedings{FSOD,
  author    = {Fan, Qi and Zhuo, Wei and Tang, Chi-Keung and Tai, Yu-Wing},
  title     = {Few-Shot Object Detection with Attention-RPN and Multi-Relation Detector},
  booktitle = {CVPR},
  year      = {2020}
}

@inproceedings{MEGA,
  author    = {Chen, Yihong and Cao, Yue and Hu, Han and Wang, Liwei},
  title     = {Memory Enhanced Global-Local Aggregation for Video Object Detection},
  booktitle = {CVPR},
  year      = {2020}
}

@inproceedings{RDN,
  author    = {Deng, Jiaqi and Pan, Yingwei and Yao, Ting and Zhou, Wengang and Li, Houqiang and Mei, Tao},
  title     = {Relation Distillation Networks for Video Object Detection},
  booktitle = {ICCV},
  year      = {2019}
}

@inproceedings{CTracker,
  author    = {Peng, Jiayuan and Wang, Chuang and Wan, Fang and Wu, Yuxuan and Wang, Yibing and Tai, Yu-Wing and Wang, Chengjie and Li, Jilin and Huang, Feiyue and Fu, Yun},
  title     = {Chained-Tracker: Chaining Paired Attentive Regression Results for End-to-End Joint Multiple-Object Detection and Tracking},
  booktitle = {ECCV},
  year      = {2020}
}

@article{FairMOT,
  author    = {Zhang, Yifu and Wang, Chunyu and Wang, Xinggang and Zeng, Weiming and Liu, Wenyu},
  title     = {FairMOT: On the Fairness of Detection and Re-Identification in Multiple Object Tracking},
  journal   = {IJCV},
  year      = {2021}
}

@inproceedings{CenterTrack,
  author    = {Zhou, Xingyi and Koltun, Vladlen and Krähenbühl, Philipp},
  title     = {Tracking Objects as Points},
  booktitle = {ECCV},
  year      = {2020}
}

@inproceedings{tacf,
  title={Temporal Aggregation with Context Focusing for Few-Shot Video Object Detection},
  author={Han, Wentao and Lei, Jie and Wang, Fahong and Feng, Zunlei and Liang, Ronghua},
  booktitle={SMC},
  year={2023},
}

@inproceedings{video-owlvit23,
  title={Video OWL-ViT: Temporally-consistent open-world localization in video},
  author={Heigold, Georg and Minderer, Matthias and Gritsenko, Alexey and Bewley, Alex and Keysers, Daniel and Lu{\v{c}}i{\'c}, Mario and Yu, Fisher and Kipf, Thomas},
  booktitle={ICCV},
  year={2023}
}

@inproceedings{hsieh2019one,
  title={One-shot object detection with co-attention and co-excitation},
  author={Hsieh, Ting-I and Lo, Yi-Chen and Chen, Hwann-Tzong and Liu, Tyng-Luh},
  booktitle={NeurIPS},
  year={2019}
}

@inproceedings{sun2021fsce,
  title={FSCE: Few-shot object detection via contrastive proposal encoding},
  author={Sun, Bo and Li, Banghuai and Cai, Shengcai and Yuan, Ye and Zhang, Chi},
  booktitle={CVPR},
  year={2021}
}

@inproceedings{dai2021up,
  title={UP-DETR: Unsupervised pre-training for object detection with transformers},
  author={Dai, Zhigang and Cai, Bolun and Lin, Yugeng and Chen, Junying},
  booktitle={CVPR},
  year={2021}
}

@inproceedings{fan2022few,
  title={Few-shot video object detection},
  author={Fan, Qi and Tang, Chi-Keung and Tai, Yu-Wing},
  booktitle={ECCV},
  year={2022},
  organization={Springer}
}

@inproceedings{zhu2017flow,
  title={Flow-guided feature aggregation for video object detection},
  author={Zhu, Xizhou and Wang, Yujie and Dai, Jifeng and Yuan, Lu and Wei, Yichen},
  booktitle={ICCV},
  year={2017}
}

@inproceedings{wu2019sequence,
  title={Sequence level semantics aggregation for video object detection},
  author={Wu, Haiping and Chen, Yuntao and Wang, Naiyan and Zhang, Zhaoxiang},
  booktitle={ICCV},
  year={2019}
}

@inproceedings{minderer2022simple,
  title={Simple open-vocabulary object detection with vision transformers},
  author={Minderer, Matthias and Gritsenko, Alexey and Stone, Austin and Neumann, Maxim and Weissenborn, Dirk and Dosovitskiy, Alexey and Mahendran, Aravindh and Arnab, Anurag and Dehghani, Mostafa and Shen, Zhuoran and Wang, Xiao and Zhai, Xiaohua and Kipf, Thomas and Houlsby, Neil},
  booktitle={ECCV},
  year={2022}
}

@inproceedings{heigold2023video,
  title={Video OWL-ViT: Temporally-consistent open-world localization in video},
  author={Heigold, Georg and Minderer, Matthias and Gritsenko, Alexey and Bewley, Alex and Keysers, Daniel and Lu{\v{c}}i{\'c}, Mario and Yu, Fisher and Kipf, Thomas},
  booktitle={ICCV},
  year={2023}
}

@inproceedings{li2022grounded,
  title={Grounded language-image pre-training},
  author={Li, Liunian Harold and Zhang, Pengchuan and Zhang, Haotian and Yang, Jianwei and Li, Chunyuan and Zhong, Yiwu and Wang, Lijuan and Yuan, Lu and Zhang, Lei and Hwang, Jenq-Neng and others},
  booktitle={CVPR},
  year={2022}
}

@inproceedings{zhang2023simple,
  title={Simple Open-Vocabulary Semantic Segmentation},
  author={Zhang, Jian and Zhao, Haobo and Wang, Binghao and Shi, Yujun and Zhu, Xiaobo and Guo, Qingyun and Cai, Tianqi and Liu, Chang and Li, Weijie and Xu, Shuang and Wang, Jingdong},
  booktitle={CVPR},
  year={2023}
}

@inproceedings{resnet,
  author = {He, Kaiming and Zhang, Xiangyu and Ren, Shaoqing and Sun, Jian},
  booktitle = {CVPR},
  title = {{Deep Residual Learning for Image Recognition}},
  year = 2016
}

@INPROCEEDINGS{llm_fsod,
  author={Han, Guangxing and Lim, Ser-Nam},
  booktitle={CVPR}, 
  title={Few-Shot Object Detection with Foundation Models}, 
  year={2024},
}

@inproceedings{mml_clas,
  title={Multi-Modal Classifiers for Open-Vocabulary Object Detection},
  author={Kaul, Prannay and Xie, Weidi and Zisserman, Andrew},
  booktitle={ICML},
  year={2023}
}

@inproceedings{tide-eccv2020,
  author    = {Daniel Bolya and Sean Foley and James Hays and Judy Hoffman},
  title     = {TIDE: A General Toolbox for Identifying Object Detection Errors},
  booktitle = {ECCV},
  year      = {2020},
}

@inproceedings{Kumar023-2,
  title = {Few-Shot Referring Relationships in Videos},
  author = {Yogesh Kumar and Anand Mishra},
  year = {2023},
  booktitle = {CVPR},
}

@inproceedings{kumar2025matr,
  title={Aligning Moments in Time using Video Queries},
  author={Yogesh Kumar and Uday Agarwal and Manish Gupta and Anand Mishra},
  booktitle={ICCV},
  year={2025},
}

@article{kumar2025moment,
  title={Moment Alignment Transformer for Video-to-Video Moment Retrieval},
  author={Kumar, Yogesh and Agarwal, Uday and Gupta, Manish and Mishra, Anand},
  journal={Authorea Preprints},
  year={2025},
  publisher={Authorea}
}

@InProceedings{Jung_2025_WACV,
    author    = {Jung, Minjoon and Jang, Youwon and Choi, Seongho and Kim, Joochan and Kim, Jin-Hwa and Zhang, Byoung-Tak},
    title     = {Background-Aware Moment Detection for Video Moment Retrieval},
    booktitle = {WACV},
    year      = {2025},
}

\end{document}